\documentclass[preprint,journal]{vgtc}       

\ifpdf
  \pdfoutput=1\relax                   
  \usepackage{graphicx}                
  \DeclareGraphicsExtensions{.pdf,.png,.jpg,.jpeg} 
\else
  \usepackage{graphicx}                
  \DeclareGraphicsExtensions{.eps}     
\fi%

\graphicspath{{figures/}{pictures/}{images/}{./}} 

\usepackage{microtype}                 
\PassOptionsToPackage{warn}{textcomp}  
\usepackage{textcomp}                  
\usepackage{mathptmx}                  
\usepackage{times}                     
\usepackage{cite}                      
\usepackage{tabu}                      
\usepackage{booktabs}                  
\usepackage{algorithm}
\usepackage[noend]{algpseudocode}
\usepackage{tikz}
\usepackage{amssymb}
\usepackage{tabularx}
\usepackage{multicol}
\usepackage{multirow}
\newcommand{\Cross}{\mathbin{\tikz [x=1.4ex,y=1.4ex,line width=.2ex] \draw (0,0) -- (1,1) (0,1) -- (1,0);}}

\usepackage{color}

\newcommand{\name}[1]{NAS-Navigator}

\onlineid{1626}

\vgtccategory{VIS 2022 Analytics \& Decisions Papers}
\vgtcpapertype{algorithm/technique}

\title{\name{}: Visual Steering for Explainable One-Shot Deep Neural Network Synthesis}

\author{Anjul Tyagi, Cong Xie, Klaus Mueller}
\authorfooter{
\item
 Anjul Tyagi, Cong Xie and Klaus Mueller are with the Visual Analytics and Imaging Lab at Computer Science Department, Stony Brook University, New York. E-mail: \{aktyagi, coxie, mueller\}@cs.stonybrook.edu.
}

\shortauthortitle{Tyagi \MakeLowercase{\textit{et al.}}: \name{}.}

\abstract{
The success of DL can be attributed to hours of parameter and architecture tuning by human experts. Neural Architecture Search (NAS) techniques aim to solve this problem by automating the search procedure for DNN architectures making it possible for non-experts to work with DNNs. Specifically, One-shot NAS techniques have recently gained popularity as they are known to reduce the search time for NAS techniques. One-Shot NAS works by training a large template network through parameter sharing which includes all the candidate NNs. This is followed by applying a procedure to rank its components through evaluating the possible candidate architectures chosen randomly. However, as these search models become increasingly powerful and diverse, they become harder to understand. Consequently, even though the search results work well, it is hard to identify search biases and control the search progression, hence a need for explainability and human-in-the-loop (HIL) One-Shot NAS. To alleviate these problems, we present \name{}, a visual analytics (VA) system aiming to solve three problems with One-Shot NAS; explainability, HIL design, and performance improvements compared to existing state-of-the-art (SOTA) techniques. \name{} gives full control of NAS back in the hands of the users while still keeping the perks of automated search, thus assisting non-expert users. Analysts can use their domain knowledge aided by cues from the interface to guide the search. Evaluation results confirm the performance of our improved One-Shot NAS algorithm is comparable to other SOTA techniques. While adding Visual Analytics (VA) using \name{} shows further improvements in search time and performance. We designed our interface in collaboration with several deep learning researchers and evaluated \name{} through a control experiment and expert interviews. 
} 


\CCScatlist{ 
 \CCScat{K.6.1}{Management of Computing and Information Systems}%
{Project and People Management}{Life Cycle};
 \CCScat{K.7.m}{The Computing Profession}{Miscellaneous}{Ethics}
}

\teaser{
  \centering
  \includegraphics[width=\linewidth]{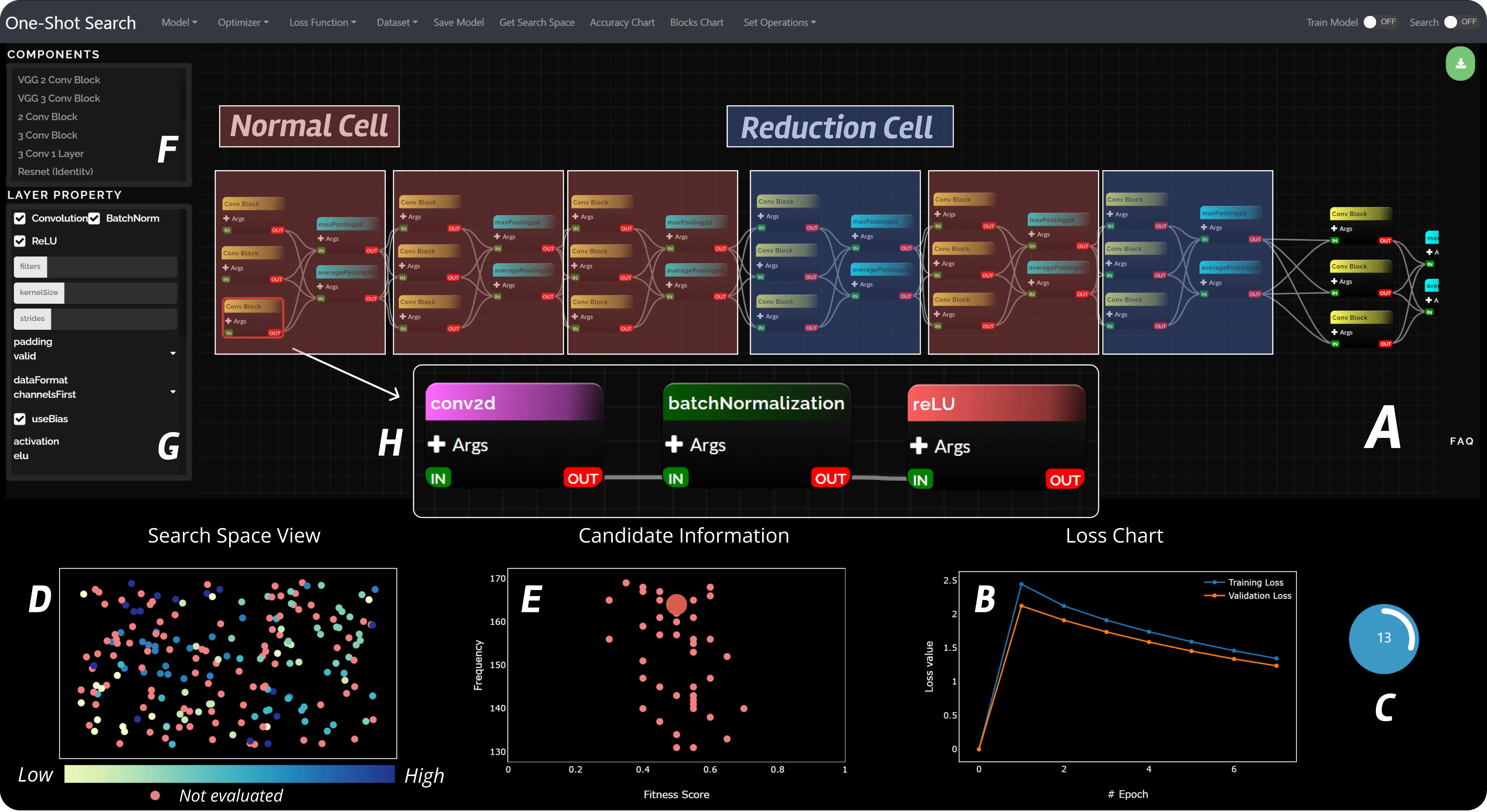} \caption{
\name{} is visual analytics (VA) framework for explainable and human-in-the-loop neural network architecture search (NAS). \name{} implements a one-shot NAS, using an iterative evolutionary search algorithm. The interface supports the visualization of NAS with the human-in-the-loop search control. Analysts start by designing a large template network, through a lego view (A); capable of emulating the search space of candidate neural networks. This network is then trained for a few epochs to initialize meaningful weights, useful for candidate NN search, via a loss chart view (B). Following this, our evolutionary search algorithm evaluates possible candidate NNs iteratively, with iteration counter (C); sampled from the large NN, and these accuracy evaluation results are then presented in the form of a candidate NN projection on a scatterplot, via a search space view (D). Analysts can further pause/stop the search and edit the template NN based on the fitness scores generated by our search algorithm, on the candidate information view (E); to generate the final NN architecture, or to reduce the search space size. The fitness scores are calculated for each node of the candidate neural networks which are sampled from the large template network during the search.}
  \label{fig:teaser}
}

\vgtcinsertpkg

\begin{document}
\maketitle

    \section{Introduction}
    With the recent advances in computing power, deep learning (DL) has made it possible to automate the problem of feature engineering through neural networks (NNs). Highly complex features can be automatically learned from the data. However, this requires carefully designed DNN architectures, which transform the problem of feature engineering to architecture engineering \cite{elsken2018neural}. Some well-studied DNNs like AlexNet~\cite{krizhevsky2012imagenet} and ResNet~\cite{he2016deep} have been the results of extensive architecture search studies and required many hours of manual parameter tuning by experts. Most of the current automated approaches find the optimal solution of a NN architecture based on adaptive experiments~\cite{pham2018efficient}, and most of them rely on strong computing power. As a result, these networks are hard to generalize because of the very high hardware equipment demands and associated costs. Network Architecture Search (NAS) techniques aim at alleviating these problems for deep learning researchers by automatically finding the best candidate NN architectures based on validation accuracy. NAS designed methods have outperformed manually curated networks as shown by Zoph \textit{et al.}~\cite{pham2018efficient}, Real \textit{et. al}~\cite{real2019regularized} and the SMASH model~\cite{brock2017smash}.
    
    Typical NAS algorithms apply techniques like Reinforcement Learning (RL) \cite{zoph_learning_2018,zoph2016neural, xu_renasrelativistic_2021} or Evolutionary Search (EA) \cite{liu_hierarchical_2018, real2019regularized} to search for candidate NNs directly. However, these approaches have been shown to be computationally very expensive. 
    One-Shot NAS techniques aim at reducing the search time for NAS through training candidate neural networks via weight sharing. The idea is that instead of training each candidate NN separately, one trains a large template NN which is a super-set network of all candidates, and then uses the same weights to randomly sample candidate NN from this main network. 

    Most NAS algorithms use Recurrent NNs (RNNs) or DNNs as the backbone to run the search. These types of algorithms typically have a low capacity for explaining their actions and strategies \cite{elsken2018neural, adadi2018peeking}. Explainability, however, is critical when deploying real-world systems which have a high need for process auditing and often also have high legal liability \cite{singh_exs_2019}. In this work, we talk about explainability from the context of NAS, and \name{} focuses on adding explainability to the one-shot NAS process. Possible benefits of this procedure include the reasoning behind why a particular NN architecture was chosen over the others. Users can also compare how different candidate NNs behave on the data. 
    Human-in-the-loop (HIL) assistance in NAS approaches can influence the choice of search in high stake decision making \cite{gil_towards_2019} and so assist human users in building trust into the process. The need for explainability and HIL is crucial for many situations and has been the subject of a long debate in the HCI community, commonly referred to as the \textit{control and automation trade-off}~\cite{horvitz_principles_1999, heer_agency_2019, shneiderman_direct_1997, amershi_guidelines_2019, tyagi2022infographics, tyagi2020ice, cao2019graphs}.
    
    Evolving from a formative study done with deep learning and NAS researchers, we developed \name{} to solve the issues revolving around these problems. \name{} is a visual analytics (VA) system that reconciles both automation and user control for NAS, where expert knowledge and automated intelligent services can be combined effectively. We present a One-Shot NAS algorithm developed using evolutionary search (EA) to support our explainability and HIL use cases. Evaluation results show that our EA algorithm is fast and more effective than typical One-Shot NAS algorithms. It learns to sample better candidates given the history of selected candidates and their validation accuracy data. We find that our scheme performs better than the random sampling strategy used in the existing One-Shot NAS techniques.
    Overall, our contributions include solutions to three problems with existing NAS techniques: 
    \begin{itemize}
    \setlength{\itemsep}{-2pt}
        \item \textbf{One-Shot NAS search speed:} Our One-Shot EA NAS algorithm provides faster search results
        \item \textbf{Explainability:} Our VA interface \name{} supports search tracking and progress, search space visualization, candidate ranking, and score visualizations to provide cues to the users
        \item \textbf{Human-in-the-loop control:} \name{} provides a user-controllable HIL NAS paradigm, where users can improve the search through VA. Users can control the final NN architecture depending on the resource-accuracy trade-off
    \end{itemize}
    
    Our evaluation through a user study and expert interviews show that \name{} is effective in adding explainability and HIL to NAS. We separately evaluate our EA One-Shot NAS algorithm for its efficiency against SOTA methods. The results we obtained show better search convergence at a similar accuracy to the final candidate model compared to existing fully automated NAS techniques. 
    
     Our paper is organized as follows. Section 2 presents related work. Section 3 describes our formative inquiry with deep learning researchers and practitioners. Section 4 presents our Explainable One-Shot NAS method. Section 5 describes our visual interface. Section 6 presents our user study and its findings. Section 7 ends with conclusions. 
     
\section{Related Work}
\begin{table}[tb]
  \caption{Comparison of \name{} with different NAS algorithms based on five aspects. \textit{VA} stands for Visual Analytics, showing whether the technique supports HIL interaction with NAS. \textit{Eff} shows if the search algorithm is efficient, i.e. the search time is less than 5 GPU days for a CNN. \textit{Sh} stands for shared parameters, meaning whether the algorithm supports search through a shared parameter template network. \textit{Exp} means explainable, comparing algorithms that support explainability in NAS. \textit{No PT} means no pre-training of the candidate networks is required for the search algorithm. The table is divided into 4 sections of rows, separating the manually designed DNNs, automated NAS techniques, VA techniques for NAS, and our work (\name{}).  
  \label{tab:comp_hyp}}
  \scriptsize%
	\centering%
  \begin{tabu}{%
	r%
	*{6}{c}%
	*{12}{r}%
	}
  \toprule
  \textbf{Method} & \textbf{VA} & \textbf{Eff} & \textbf{Sh} & \textbf{Exp} & \textbf{No PT}\\
  \midrule
  ResNet~\cite{he2016deep} & $\Cross$ & $\Cross$ & - & $\Cross$ & \checkmark \\
  \midrule
  NASNet~\cite{zoph_learning_2018} & $\Cross$ & $\Cross$ & $\Cross$ & $\Cross$ & $\Cross$ \\
  EAS~\cite{pham2018efficient} & $\Cross$ & \checkmark & \checkmark & $\Cross$ & $\Cross$ \\
  PNAS~\cite{liu_progressive_2018} & $\Cross$ & $\Cross$ & $\Cross$ & $\Cross$ & $\Cross$ \\
  SMASH~\cite{brock2017smash} & $\Cross$ & \checkmark & $\Cross$ & $\Cross$ & \checkmark \\
  DARTS~\cite{liu2018darts} & $\Cross$ & \checkmark & \checkmark & $\Cross$ & \checkmark \\
  SETN~\cite{dong_one-shot_2019} & $\Cross$ & \checkmark & \checkmark & $\Cross$ & \checkmark \\
  \midrule
  BEAMES~\cite{das2019beames} & \checkmark & - & - & \checkmark & $\Cross$ \\
  REMAP~\cite{cashman2019ablate} & \checkmark & \checkmark & $\Cross$ & \checkmark & $\Cross$ \\
  TREEPOD~\cite{muhlbacher2017treepod} & \checkmark & - & - & \checkmark & $\Cross$ \\
  \midrule
  \textbf{\name{}} & \checkmark & \checkmark & \checkmark & \checkmark & \checkmark \\
 \bottomrule
  \end{tabu}%
\end{table}

\label{s:related_work}
We summarize several NAS research works by comparing them based on five factors as shown in Table \ref{tab:comp_hyp}. Overall, these methods can be divided into 3 categories; automated NAS, One-Shot NAS, and HIL NAS. 

\textbf{Automated NAS.} Automated NAS has a long history~\cite{miller1989designing, schmidhuber1987evolutionary}. NAS designed methods have outperformed manually curated networks as shown by Zoph \textit{et al.}~\cite{pham2018efficient}, Real \textit{et al.}~\cite{real2019regularized} and the SMASH model~\cite{brock2017smash}. They use several trained networks to provide the final architecture design after evaluating each network on a validation set. DARTS~\cite{liu2018darts} is another popular NAS algorithm that searches for good candidate NNs through gradient descent. However, training networks with NAS is expensive since many different networks have to be trained before evaluation. Also, these methods lack the human-in-the-loop control and visual analytics, supporting explainability, as shown in Table \ref{tab:comp_hyp}. To overcome this, another technique called the MorphNet~\cite{gordon2018morphnet} uses a different approach where the final architecture design is decided directly as a subset of a single hypernetwork where the candidate NNs share the same parameters. Liu \textit{et al.}~\cite{liu_hierarchical_2018} proposed an evolutionary search algorithm for automated NAS without the weight sharing network search method. Evolutionary search is used to generate model architectures by manipulating operations and editing edges in the network.

\textbf{One-Shot NAS. }Following the work on MorphNets, a slightly different approach known as One-Shot Architecture Search~\cite{bender2018understanding} has evolved, which involves searching for the best neural network architecture as a subset of a largely trained hypernetwork. The hypernetwork in One-Shot NAS has a lesser number of parameters than training several different architectures independently~\cite{bender2018understanding}. Our EA algorithm used in \name{} is similar to previous one-shot approaches~\cite{zhang_graph_2020, brock2017smash, dong_one-shot_2019} where we train a hypernetwork to generate representative weights for every network in the search space (shared parameters). Although these algorithms are less resource-intensive than typical NAS algorithms, they still lack in the explainability and HIL aspects. Our EA algorithm is developed to sample optimal NN candidates along with explainability and HIL support to the typical one-shot NAS pipeline. Real \textit{et al.}~\cite{real2019regularized} proposed a one-shot algorithm specifically developed to generate a model AmoebaNet-A for hand sketches. The model performance is evaluated after generation by separate training. This is different in \name{} where we refer to that information from the template network and hence it is faster. The child models are generated by mutating the NNs from the highest accuracy models, whereas in our algorithm, children are generated from blocks with the highest fitness scores. Fitness scores include the history of that block and how it performed in all the previous models.

\textbf{Interactive NAS.}
There has been significant research in the visualization community to make NN model selection and search more effective. Techniques exist to support NAS where the model parameters are known, and the model has to be evaluated only with a single dataset~\cite{zhang2018manifold, cashman2019user, muhlbacher2013partition, schneider2018integrating}. VA frameworks have also been proposed for HIL ML applications~\cite{kumar2019task, tyagi2022visualization}. BEAMES~\cite{das2019beames} helps the users to find the best regression models for a given dataset iteratively. TreePOD~\cite{muhlbacher2017treepod} provides an interface to manage the trade-off between accuracy and interpretability of different existing ML models. REMAP~\cite{cashman2019ablate} allows interactive CNN NAS starting with a few pre-trained models. Besides designing NNs, other tools allow interactive design and filtering of clustering techniques~\cite{cavallo2018clustrophile, kwon2017clustervision, nam2007clustersculptor, sacha2017somflow} and dimension reduction~\cite{anand2012visual, choo2013interactive, jeong2009ipca, liu2015visual, nam2012tripadvisor}. However, it is considerably harder to interactively optimize a DNN compared to optimizing a regression, clustering, or dimension reduction model; \name{} contributes by adding a HIL VA interface to one-shot NAS.    

Out of all these existing techniques, the work by Cashman \textit{et al.}~\cite{cashman2019ablate}, REMAP is the most closely related to \name{}. However, REMAP does a global search by evaluating some set of models in a given search space where the accuracies of each model on a given dataset are already known. Getting this initial data where accuracy information is known is resource-intensive, and is not easily available for different applications. Datasets like NAS benchmarks ~\cite{ying2019bench, dong2020bench, siems2020bench} are not available for most applications where deep learning is applied. Hence, in \name{}, using the one-shot technique, there is no requirement for an initial dataset where every candidate is pre-evaluated. 

Another variation of \name{} from REMAP is in the ablation and variation phases. REMAP provides the user with options to drop some layers in the original NN for evaluation of different connections in the NN search space. In \name{}, we implement this operation automatically using the dropout operation during the search procedure. Also, users can still edit layers manually using the lego-view. Similarly, in the variation step of REMAP, users generate variations of different components of a candidate NN. \name{} handles that automatically through repeated components in the template network. Each layer in the template network consists of differently parameterized components. And the search procedure searches through all these combinations automatically. 

Besides, there are other features in \name{} which are not available in existing works. The fitness scores of each block visualization allow the users to see which regions in the search space are impactful. For networks with skip connections, the lego view provides better visualization of the architecture. Also, users can easily set parameters using the parameter visualization sidebar, which can be easily viewed for each block. The search space view provides a single view to compare the accuracy and architectural similarities of the full search space.

\section{Formative Study}
\label{s:design_study}
To systematically evolve our idea of an explainable and HIL NAS framework, we first conducted a formative study to collect requirements from deep learning researchers, their views of explainable NAS, and general workflows. This approach helped concertize our framework and tool design with a user-centered evaluation at an earlier development stage. 

The formative study participants were carefully chosen to be data analysts and researchers with different experiences, working in deep learning applications with a basic understanding of NAS techniques. Out of ten participants, two were deep learning professionals working in the industry, two were professors working in computer vision and NLP, and three were Ph.D. students working in computer vision and NLP, categorized as experts for this study (E1-E7). Three were graduate students in Computer Science studying deep learning with a basic understanding of NAS, categorized as non-experts for this study (NE1-NE3). Each participant was interviewed for about 45 minutes discussing their experience in DNNs and NAS. We covered the following topics during the interview, categorized as who proposed the design ideas.

\begin{itemize}
\setlength{\itemsep}{-2pt}
    \item Their experience with the general NAS and One-Shot NAS workflows.
    \item Principles, practices, and difficulties of NAS.
    \item Benefits and frustrations of existing VA machine learning tools.
\end{itemize}



\subsection{Key Findings - Design Components}
\label{ss:design_components}
The purpose of the formative study was to gather a list of requirements from domain experts and potential users, which are expected to be met by our framework. Our many discussions culminated in the following list of requirements.
\begin{itemize}
\setlength{\itemsep}{-2pt}
\item \textbf{T1: HIL One-Shot NAS search.} Develop a one-shot NAS algorithm that can be controlled through user feedback. Complete transparency on how the algorithm is searching through different NN candidates is important for explainability. Users should be given control of the search process via a VA interface. \textit{E1, E3, NE1, NE2, NE3.}

\item \textbf{T2: Template models VA.} The design of hypernetworks (template NNs) plays a crucial role in the results of One-Shot NAS techniques. Through \name{}, users should be able to design and edit template models, make choices based on their experience and search progression feedback. \textit{E2, E4, E5, NE2.}

\item \textbf{T3: Candidate models VA.} Through \name{}, users should be able to see which parts of the template NN combine to form a candidate NN for search. The interface should allow for comparing and contrasting different candidate NNs. \textit{E3, E4, E6, E7, NE1.}

\item \textbf{T4: Search Space VA.} Users should be able to see a visualization of the candidate NNs search space. This is useful to compare how different candidates are related to each other. Some cues to validate candidate performance during search evaluation would be useful. This can be used to cluster the search into different regions based on validation accuracy. \textit{E4, E7, NE3.}
\end{itemize}
\section{Our Explainable One-Shot NAS Methodology}
\label{ss:background}



In this section, we discuss the technical details of designing a One-Shot NAS technique. NAS problems are often confined to predicting the structure of different subsets of a large template NN (known as cells) instead of the complete architecture designs. This strategy is shown to be more effective than finding complete architectures of DNNs~\cite{liu_progressive_2018, liu2018darts, simonyan2014very, zhang_graph_2020}. These cells can be combined via an evaluation strategy to form a complete DNN structure, where NAS aims to find the structure of each of these cells in a DNN. A cell (in the context of CNNs) is a fully convolutional structure that maps an input tensor to an output tensor. Following the previous NAS works~\cite{liu_hierarchical_2018, simonyan2013deep}, two types of cell structures have been found useful in the context of designing CNNs, i.e. Normal Cells and Reduction Cells. 

A normal cell has CNN components with a stride of 1, which maps the input size feature map of a given height, width, and the number of feature maps (H, W, F) to the same size output feature maps (H', W', F'). A reduction cell is used to reduce the height and width of the input feature map by a factor of 2, hence (H', W', F') = (H/2, W/2, 2F). 

Each cell contains B number of nodes, with a default value of 4, they can be changed through the interface. Each node is connected to every other node through six operations (O's). These values are kept similar to past works in NAS~\cite{dong_one-shot_2019,liu_hierarchical_2018} and are: \textit{3x3 Max Pooling, 3x3 Average Pooling, Skip connection, 3x3 Separable Conv, 5x5 Separable Conv, and 1x3 then 3x1 Conv.} Each node inside a cell takes two inputs ($I_{1}$, $I_{2}$) and returns a transformed tensor $T_{o} = t_{1}(I_{1}) + t_{2}(I_{2})$ where \textit{t} refers to the transformed input tensor through an operation \textit{O}. The task is to find which of these O's work best for every pair of connected nodes for every cell.

Once the cells are identified, the overall structure of the CNNs is created using the structures identified in~\cite{zoph_learning_2018}. In a typical NAS algorithm, the cell structures are fixed and users cannot change or skip the search of particular cells/operations based on evaluated candidates. However, \name{} through a VA interface, allows users to visualize these structures and make changes at any stage of the search. Users can edit the number of normal and reduction cells, and edit nodes (fix, remove and add) inside cells in real-time based on search progression results being displayed on the interface.

This idea of cell-based construction has been extended to transformer models related to non computer vision tasks. Different templates for transformer models have been suggested~\cite{gao2021autobert}, which can be directly imported into \name{} to perform the search as the other computer vision model counterpart.

\subsection{Methodology}
\label{s:implementation}

\begin{figure*} [!h]
    \centering
        \includegraphics[width=\textwidth]{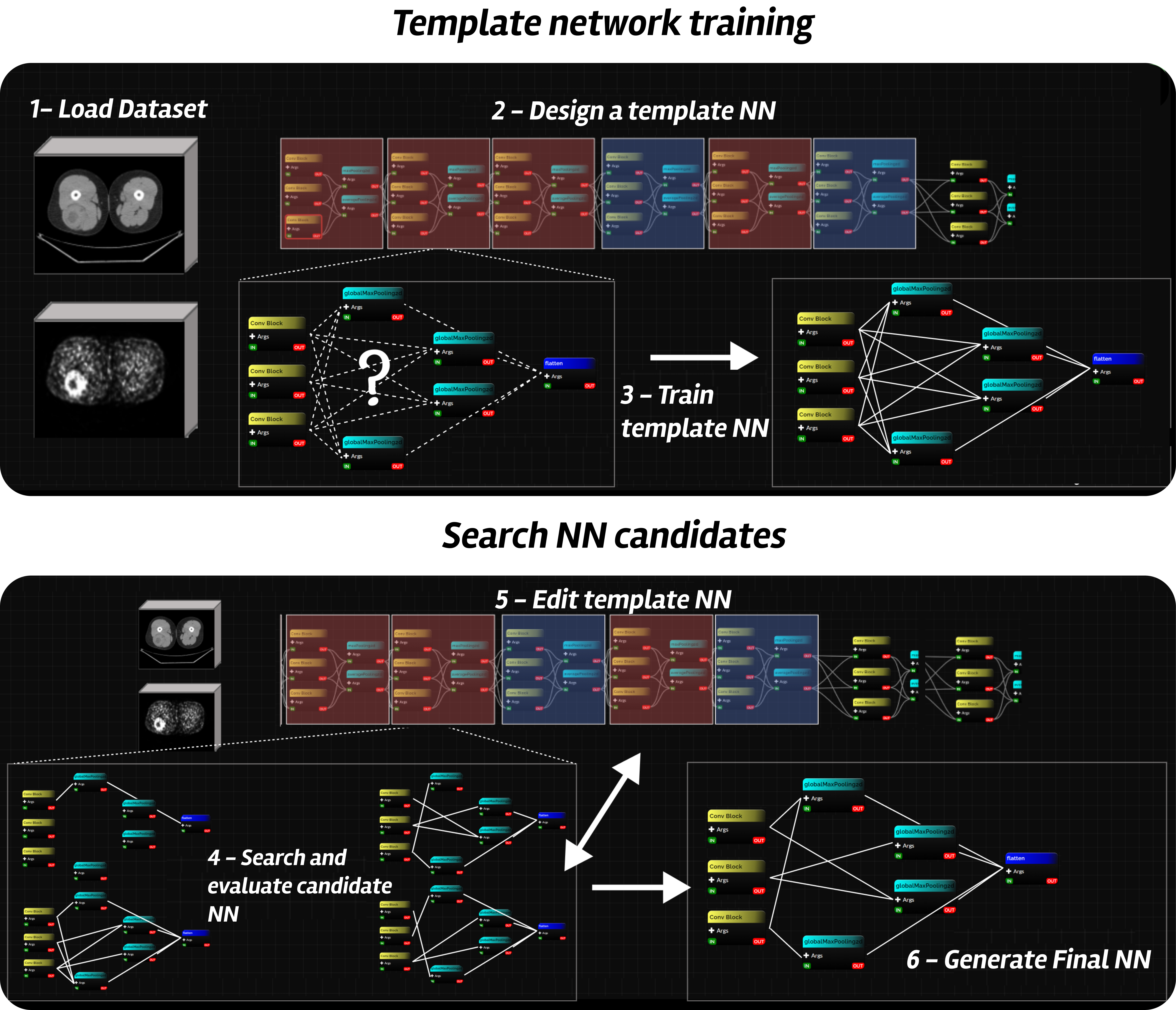}
        \vspace{-6mm}
    \caption{Stages in a one-shot NAS search process are mainly divided into two parts, template network training, and the search algorithm to find possible candidates from the trained template NN. After loading the dataset (1), a template network is trained (shown as the transition from dashed lines to solid lines in the figure) with shared parameters (2). This template network consists of multiple repeated cells (normal and reduction cells,  as described in Section~\ref{ss:background}) which are initially randomly initialized and are assigned meaningful weights after training the template network (3). Then one-shot search generates candidate NNs for evaluation from the template network (4), from which a final candidate is sampled iteratively; with the search process. During the search, users can edit the template NN using the search feedback (5); to make changes to the template NN architecture, which in-turn impacts the search progress and generation of candidate NNs (6).} 
    \label{fig:one-shot-nas}
\end{figure*}

Figure~\ref{fig:one-shot-nas} shows our methodology for implementing an explainable HIL One-Shot NAS framework with \name{}. We explain all the five stages of the process in the following text.

\textbf{S1: Dataset. }
Users can choose a dataset using the menu bar on \name{} which controls the type of template networks the user can choose through the interface. While designing a template network, the structure of the network depends on the dataset and the global structure of the DNN. For example, CNNs for CIFAR10 and ImageNet have different template structures~\cite{zoph_learning_2018}. With the help of a formative study discussed in Section~\ref{s:design_study}, we have designed template networks for 10 common datasets. 

\textbf{S2: Design a template NN (T2). }
Through \name{}, users can edit the number of normal and reduction cells in the default template network; choose the number of nodes in a cell, and change the number of cells to control the depth of the template NN. To support the tasks where the default template network is not available, users can create their template network by combining components available in the sidebar (See \textit{F} in Figure~\ref{fig:teaser}). The sidebar provides template structures for CNN and LSTM components which can be combined to create template networks. 
This satisfies the design component requirement (T2) discussed in Section~\ref{ss:design_components}.

\textbf{S3: Train the template NN (T2). }
The next step after selecting the template network structure is to train the template NN for a few epochs. This is important to assign meaningful weights to each node of the network. Users can choose when to stop the training, based on time and resource usage demands. 
The template network train accuracy affects the search results, hence more training will result in better search results, but also higher training resource consumption.  
Also, using the cues from the search iterations, users can edit the template network to add or remove nodes or cells. This satisfies the design requirement (T2) from Section~\ref{ss:design_components}.

\textbf{S4: Search Algorithm (T1, T3, T4). }
Training the template network assigns meaningful weights to each path (node and their respective operations inside a cell) in the network. 
After training, the purpose of the search algorithm is to evaluate these paths in the template NN cells and choose the operations with the highest fitness scores. This can be presented as a graph search algorithm where each candidate NN is a path in the super-graph connecting the start and end nodes. In every search iteration, a few of these subgraphs (candidate NNs) are evaluated for validation accuracy and corresponding nodes and operations contained in that candidate are ranked. This procedure ranks and helps finalize the best performing cell structure, and hence the candidate NN. Our EA-based search algorithm evaluates the candidate NNs in each iteration and learns from their validation accuracies, thus using this information to choose better candidates in the next iteration. The algorithm updates the candidate information view (E) and the search space view (D) as shown in Figure~\ref{fig:teaser}. More details on our search algorithm are discussed in Section~\ref{ss:search}. This satisfied the design requirement (T1, T3, and T4) formulated in Section~\ref{ss:design_components}.

\textbf{S5: Final Architecture Evaluation. }
The last stage of this complete procedure is to evaluate the best candidates found with the help of the search algorithm. Users can choose to stop the search at any iteration based on the intermediate results and the amount of search space explored by the algorithm. The final model consists of one of the final candidates with the highest validation accuracy.

\subsection{Evolutionary Search Algorithm. }
\label{ss:search}

\begin{figure*}[!ht]
    \centering
        \includegraphics[width=\textwidth]{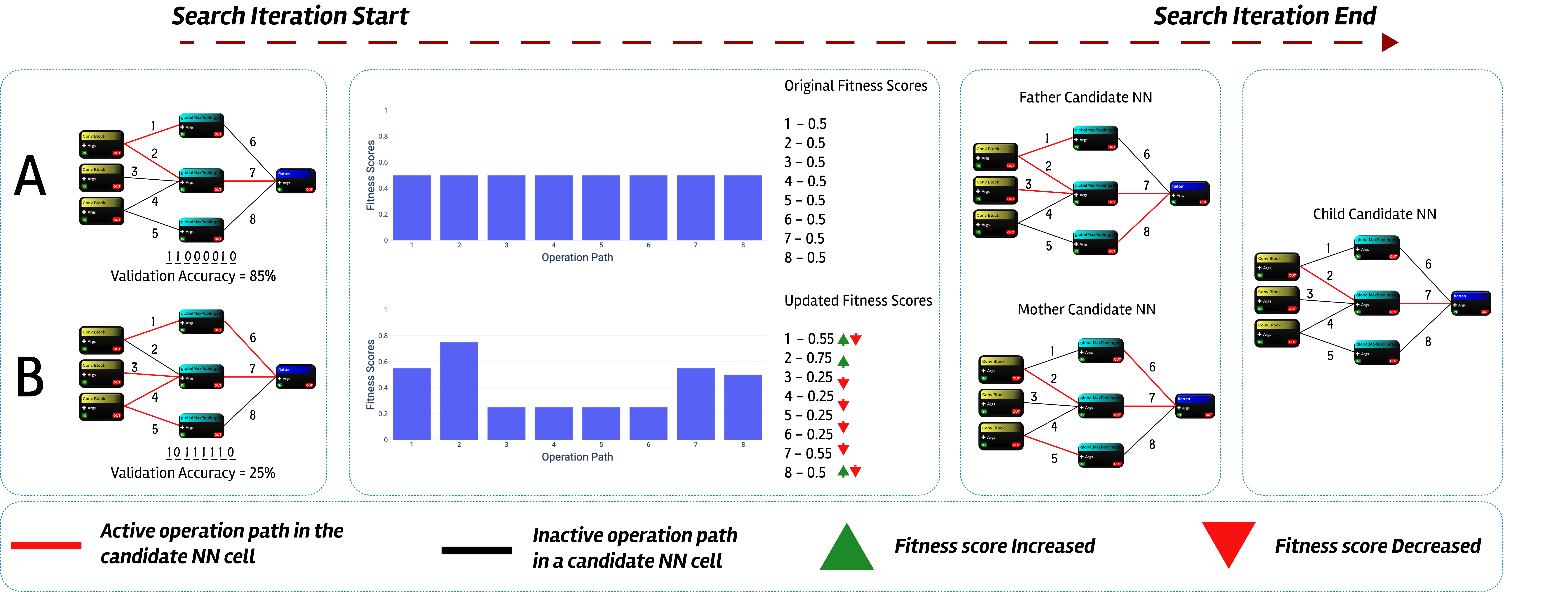}
        \vspace{-6mm}
    \caption{Our EA One-Shot NAS search algorithm overview. Assuming two candidate NNs \textit{A and B} sampled from a template NN cell, are evaluated in the current search iteration. After the validation accuracies are available for these two candidates, the fitness scores associated with each path of the template NN cell are updated. In this example, all the paths had equal fitness scores before the search iteration, which are updated after the validation of candidate NNs. Based on these new fitness scores, a father and mother candidate NNs are a sample from the distribution of fitness scores, and a child mask is generated using the cross-over algorithm. This child mask is the candidate NN for the next search iteration. This way, our EA algorithm generates more child candidates coming from higher fitness score paths. } 
    \label{fig:ea_search}
\end{figure*}

We devised an evolutionary search algorithm to search for the candidate NN architectures with HIL and add explainability to the process. The stages of our EA search algorithm are summarized in Figure~\ref{fig:ea_search} and are discussed in detail below. 

\textbf{EA1: Selecting the candidate models. } 
To select each candidate NN, we generate a bitmask vector with each bit corresponding to a path inside each cell, an example shown in Figure~\ref{fig:ea_search} for candidate NNs \textit{A} and \textit{B}. Path refers to the connection between blocks in the network. This mask represents a subgraph from the template NN which includes the paths corresponding to set bits. We generate a candidate NN from the mask by zeroing out the weights of the paths excluded from the template network. This way, only the paths with a corresponding set bit in the mask are activated. The number of candidate NNs in the population is heuristically set to 1.5 times the number of cells in the template neural network based on the studies shown in \cite{chen2015measuring}, that relate the dataset dimensionality with the population size in EA algorithms. 



\textbf{EA2: Calculating the fitness scores. }
To calculate the fitness of a candidate NN (subgraph), we calculate its validation accuracy and use Equation~\ref{eq: fitness} to alter the fitness scores of the paths existing in the candidate NN. $\alpha$ is an accuracy threshold that we fix for our experiments, which serves as a tradeoff on how fast the search algorithm learns from the current accuracy scores. Following this, all the fitness scores for a particular cell are normalized by dividing the sum of all fitness scores from individual node fitness values. 

\setlength{\abovedisplayskip}{5pt}
\begin{equation}
\label{eq: fitness}
    individualPathFitness = ValidationAccuracy - \alpha
\end{equation}

\textbf{EA3: Choose Parents. }
The choose parents procedure returns the father and the mother candidate NNs from the fitness probability distributions of the population. This means that there is a higher probability of choosing the models with better fitness scores, as shown in the second block in Figure~\ref{fig:ea_search}. 

\textbf{EA4: Cross-Over. }
To generate the child architectures from the father and mother NNs, a cross-over procedure is used, shown in Algorithm~\ref{alg:search} (Cross-Over). Firstly, a mask is generated with the procedure described in the state \textit{EA1} above. This mask is compared to the masks of the father and mother models to generate a child mask. Both the father and mother models are chosen from the best performing candidates in the population. Also, the mask is generated from the probability density of the fitness scores, i.e. more probability of bits being set at indices pointing to well-performing paths. 





\textbf{EA5: Mutation. }
The child mask is mutated as shown in Algorithm~\ref{alg:search} (Mutation), to generate a final child candidate. We chose the mutation rate to be 0.05 based on the work by Suganuma \textit{et al.}\cite{suganuma2017genetic}.


\textbf{EA6: Get Mask Probabilities. }
At each search iteration, when the new population is updated, each path in the template network is given a fitness score between 0 and 1, see part \textit{EA2} above. These values combined, form the probability distribution of the paths from which paths for the next candidate NNs are sampled. 

\textbf{Complete Evolutionary Search Algorithm. }
The complete search algorithm combines all these different stages in each iteration with a goal to find better candidates in each iteration, learning from previous candidates' performances. Algorithm~\ref{alg:search} explains our complete EA search, also shown in Figure~\ref{fig:ea_search}. 

\begin{algorithm}
\caption{One-Shot Evolutionary Search Algorithm}\label{alg:search}
\begin{algorithmic}[1]
\Procedure{crossOver}{father, mother}
\State $mask\gets$ Randomly generated mask
\State $childMask\gets (mask==0)*(father.mask) + (mask==1)*(mother.mask)$
\\
\Return $childMask$
\EndProcedure
\\
\Procedure{mutation}{newModelMask}
\State $mask\gets$ Uniformly generated numbers from 0 to 1
\State $mutationRate \gets 0.05$
\State $childMask\gets newModelMask$ AND $(mask>mutationRate)$ OR $(1-childMask)$ AND $(mask <= mutationRate)$
\State $childModel \gets getModel(childMask)$
\\
\Return $childModel$
\EndProcedure
\\
\Procedure{Architecture Search}{}
\State $population\gets$ Set of candidate models
\State $loss\gets$ Set of loss values
\For{each iteration}
\State $population, loss, maskProb \gets EVOLVE(population)$
\State $loss.append([max(loss), mean(loss), min(loss)])$
\EndFor
\EndProcedure
\\
\Procedure{Evolve}{population}
\State $fitness \gets$ fitness scores for each NN in population
\State $newPop \gets$ Top \textit{k} NN from population with highest fitness
\State $k$ \textless $length(population)$
\For{i=0 \textbf{to} length(population)-k \textbf{step} 1}
\State $father, mother \gets chooseParents(population, fitness)$
\State $newModel \gets crossOver(father, mother)$
\State $newModelMask \gets mutate(newModel)$
\State $newPop.append(newModelMask)$
\EndFor
\State $loss \gets getLoss(newPop)$
\State $maskProb \gets getMaskProb(newPop)$
\\
\Return $newPop, loss, maskProb$
\EndProcedure
\end{algorithmic}
\end{algorithm}

\section{The Interface}
\label{s:tool_description}
To implement our idea of explainable HIL NAS, we developed \name{} with the help of principles discovered during the formative study (Section~\ref{s:design_study}). \name{} allows users to interactively control the search algorithm of our framework. As shown in Figure~\ref{fig:teaser}, our interface consists of six views, which we discuss in the following text. The \textit{T\#} and \textit{S\#} next to each view show which of the formative study requirement and the One-Shot NAS stage that view satisfies.

\subsection{Lego View (T1, T2, T3, S2, S4)}
Shown in Figure~\ref{fig:teaser}(A), the goal of the Lego view is to allow editing of the template NN. Users can control the network depth, edit different components and visualize how the components are placed in this view. Controlling the template NN gives the power to the users to control the search. Also, users can visualize candidate NNs, which are a subgraph of the template NN in this view, where the cells and the nodes contained inside a candidate NN can be highlighted over the template NN.   

\subsection{Loss Chart (T1, T2, S3, S5)}
As shown in Figure~\ref{fig:teaser} (B), the loss chart allows users to visualize and control the training of template NN before running our EA search algorithm. As discussed in Section~\ref{s:implementation} (S3), training of the template NN is crucial to assign meaningful weights to the nodes. Users can make choices on how much to train the template NN based on this loss chart and depending on the resource availability for training. We can also view the final evaluation of the selected candidate NN in this view. 

\subsection{Search Space View (T1, T4, S4)}
\label{ss:ssv}
As shown in Figure~\ref{fig:teaser} (D), the search space view is a scatterplot projection of candidate NN space. This is one of the key components of \name{} since it allows users to visualize and interact directly with the candidate NN space. These projections are obtained using t-SNE~\cite{maaten2008visualizing} and graph edit distance~\cite{abu2015exact} on a randomly sampled set of candidate NNs. The sampled candidate NNs are subgraphs of the template NN with nodes labeled by the component or operation type, e.g. C for convolution and R for Relu. Using these labeled directed subgraphs, the distances between each of the sampled candidate architectures are calculated using the graph edit distance which is stored in a distance matrix. A t-SNE projection is generated from this distance matrix in 2-D shown as the search space view. This clusters the search space based on the architectural similarity of the candidate models. As the search algorithm progresses, these candidates are colored based on their evaluation accuracy. Hence, the search space view acts as a dual clustered space for architecture and accuracy similarity. 

There are several user interactions supported in the search space view. Hovering over each dot highlights the candidate NN in the lego view. As the search iterations progress, the evaluated candidate NNs are colored based on their evaluation scores, see Figure~\ref{fig:teaser} (D). This is useful to cluster the candidate search space, as more search iterations (shown as C in Figure~\ref{fig:teaser}) are elapsed, the search space view will be clustered based on the candidate performances. This can help the users to separate the search space into high-scoring and low-scoring candidate regions. Using this information, users can also select a region in the search space, which then limits the search algorithm to select candidates from the selected region for further iterations. 

Another useful piece of information presented with the search space view is the set operations on nodes and operations. Users can drag areas on the scatterplot and find the \textit{Union, Intersection or Complement} of the nodes and operations in that region. This is a helpful operation to find the most useful components of the search space which allows users to edit the template NN, thus impacting the search algorithm directly. For example, users can remove the most common cells (intersection) from a low-scoring search space region from the template NN, which reduces the search space, thus allowing for faster convergence of the EA algorithm. 

\subsection{Candidate Information View (T1, T3, S4)}
As shown in Figure~\ref{fig:teaser} (E), the candidate information view is a scatterplot showing the relationship between frequency and fitness scores of paths in the template NN. The frequency shows how many candidate NNs contain a particular path, hence more frequency score means higher occurrence. The fitness scores are assigned during each search iteration. 

The idea behind the candidate information view is to show confidence in the fitness scores of the paths. For example, removing low-scoring, high-frequency paths from the template NN can prune the search space and help converge the EA algorithm faster. Also, this view helps analyze each operation path in the template NN and their respective fitness scores which have been accumulated over the search progression. Hovering over an operation in the lego view highlights the corresponding dot on the candidate information view. Similarly, dragging an area on the scatterplot highlights all the paths contained in that area on the lego view. 

\subsection{Menu Bar, Properties Sidebar (T2, S1, S2)}
Shown in Figure~\ref{fig:teaser} (F,G), the menu bar provides buttons for selecting the model, optimizers, loss functions, datasets, saving a model, and set operations (discussed in Section~\ref{ss:ssv}). The properties sidebar is linked with the lego view and displays the properties of each node and operation in the template NN. Users can change the parameters for each node through this sidebar. 


\section{Evaluation}
\label{s:evaluation}
In this section, we evaluate \name{} for its effectiveness and design efficiency through a fully automated One-Shot NAS comparison with the state-of-the-art (SOTA), followed by case studies to show how HIL and VA can support better NAS. Finally, we evaluate the design experience of using \name{} through a user study and expert user interviews. 

\subsection{Comparison of SOTA and our EA One-Shot NAS algorithms}
In this study, we evaluate the performance of our EA algorithm against the existing NAS techniques on ImageNet~\cite{5206848}. ImageNet is a large-scale image classification dataset that has been extensively used for evaluating computer vision object detection research. The dataset contains 1.28 million training images with 50k validation images. For this study, we ran our EA algorithm for 1k iterations and experimented with different fitness-scores threshold values to create our final NN. Based on our experiments, the fitness-score threshold of 0.68 was used to obtain maximum accuracy on ImageNet with our model. We train our template network with a batch size of 256 for 400 epochs using an SGD optimizer. We use the setting similar to previous NAS training methods~\cite{dong_one-shot_2019} for setting the learning rate to 0.025 and decay to zero using the cosine scheduler. The probability of dropout for our EA algorithm was set to 0.1. Table~\ref{tab:study1} shows the comparison of our algorithm against the existing techniques. We separate the existing techniques based on their efficiency of NAS along with human-derived NNs, shown under \textit{Task Category}.  

As shown in the results, Table~\ref{tab:study1}, our model obtained the best Top-5 Acc, the same as the previous best performing model GDAS~\cite{dong2019searching} (best of 5 experiment runs). Other parameters for our model are comparable to the existing NAS techniques. Our search algorithm took about 1.7 GPU days on a Tesla V100 GPU to run for 200 iterations. 



\begin{table*}[tb]
  \centering
  \caption{Comparing our EA algorithm to existing State-of-the-art NAS techniques and human designed CNNs on ImageNet. \textit{Task Category} groups the models based on their NAS effectiveness, separately placing the human designed CNNs. We run our EA algorithm on Testla V100 GPUs for 200 iterations, achieving best Top-5 Acc with our experiments, similar to GDAS. The experiments took approximately 5 hrs on \name{} with 8 GPUs.}
    \begin{tabular}{cp{10.635em}cccc}
    \toprule
          \textbf{Task Category} & \textbf{Method} & \multicolumn{1}{p{4.045em}}{\textbf{GPU days}} & \multicolumn{1}{p{7.68em}}{\textbf{Parameters (MB)}} & \multicolumn{1}{p{5.045em}}{\textbf{Top-1 Acc}} & \multicolumn{1}{p{6.135em}}{\textbf{Top-5 Acc}} \\
    \midrule
    \multicolumn{1}{c}{\multirow{2}[4]{*}{Human Experts}} & ResNet~\cite{he2016deep} & \multicolumn{1}{p{4.045em}}{-} & 11.7  & 69.8  & 89.1 \\
\cmidrule{2-6}          & Inception-v1~\cite{al2017deep} & \multicolumn{1}{p{4.045em}}{-} & 6.6   & 69.8  & 89.9 \\
    \midrule
    \multicolumn{1}{c}{\multirow{2}[4]{*}{NAS with more than 100 GPU days}} & Progressive NAS~\cite{liu_progressive_2018} & 150   & 5.1   & \textbf{74.2} & 91.9 \\
\cmidrule{2-6}          & NASNet~\cite{qin2019nasnet} & 2000  & 5.3   & 72.8  & 91.3 \\
    \midrule
    \multicolumn{1}{c}{\multirow{5}[10]{*}{NAS with less than 5 GPU days}} & DARTS~\cite{liu2018darts} & 4     & 4.9   & 73.1  & 91 \\
\cmidrule{2-6}          & SNAS~\cite{xie_snas_2018}  & \textbf{1.4} & \textbf{4.3} & 72.7  & 90.8 \\
\cmidrule{2-6}          & \textbf{GDAS}~\cite{dong2019searching}  & 0.85  & 5.3   & 74    & \textbf{91.5} \\
\cmidrule{2-6}          & SETN~\cite{dong_one-shot_2019}  & 1.8   & 5.3   & 74.1  & 91.4 \\
\cmidrule{2-6}          & \textbf{\name{} (T=200)} & 1.7   & 5.3   & 73.9  & \textbf{91.5} \\
    \bottomrule
    \end{tabular}%
  \label{tab:study1}%
\end{table*}%

\subsection{Case Studies}
\label{ss:case_study}
Compared to a fully automated study, we separately did case studies with real users to see the impact of HIL and VA on One-Shot NAS. The goal of this study was to compare the efficiency of our framework for search time, resource usage, and usability. We set a baseline accuracy range for our experiments which was devised based on the best performing existing NAS techniques and our automated EA algorithm.  For each experiment, we noted the amount of GPU days it requires for our users to achieve that accuracy through our system. For this study, we used the CIFAR10 and CIFAR100 datasets~\cite{krizhevsky2009learning} containing 60k images categorized into 10 and 100 categories for object detection.

\textbf{Participants.} We worked with six participants for this study, who were chosen based on their experience levels with Deep Learning on Computer Vision tasks with NAS techniques. 3 were categorized as experts and 3 as Non-experts according to their experience, with experts having experience working with NAS and DL for more than 3 years. The experts were Ph.D. students in Computer Vision and non-experts were graduate students working in Computer Science with basic knowledge of Deep Learning and Computer Vision. Out of the six participants, 4 were males and 2 were females. 2 experts and 2 non-experts were the users from the formative study, who proposed the design of the original system (See Sec~\ref{s:design_study}). 3 experts used in the further evaluation are listed as E1-E3. 

\textbf{Task Description. }We initially informed the participants of the concepts of NAS and related terminologies for our framework. Next, we showed them a few examples from the template networks and the filtering steps possible with our interface. The participants were then allowed to experiment with the framework and ask clarifying questions regarding the tasks. The task for the participants in this study was to achieve a baseline accuracy range on both the CIFAR datasets. The accuracy values were decided based on existing NAS techniques and CNN models evaluated on these datasets, as shown in Table~\ref{tab:study2}. To get the accuracy baselines for the case study tasks, we used the previous works from Table~\ref{tab:comp_hyp} and Table~\ref{tab:study1} which have published results on CIFAR10 and CIFAR100 datasets, separated into three categories of Human Experts, Older NAS techniques taking $>$ 100 GPU days and newer one-shot techniques taking $<$ 5 GPU days. The task for the users was to use \name{} and based on their knowledge and cues provided in the interface, get the accuracy results within the baseline range. The time taken for our expert and non-expert users to achieve this task is reported in Table~\ref{tab:study2}.

\textbf{Results.} Table~\ref{tab:study2} shows the comparison of time and accuracies achieved by \name{} with Experts and non-expert users. We can see that using \name{} both the expert and non-expert users achieved the desired accuracies in 0.6 to 1.5 GPU days. For expert users, using domain knowledge and search pruning helped achieve the accuracies in 0.6-0.9 GPU days, which is considerably less than all the existing NAS techniques. This evaluates the efficacy of VA and HIL in NAS and the impact of domain knowledge in reducing the search times. Even for non-expert users, the time taken is less than our fully automated EA algorithm, achieving similar accuracies. 

\begin{table*}[tb]
  \centering
  \caption{Evaluating our HIL One-Shot NAS framework with fully automated NAS techniques. We categorize existing NAS techniques and CNNs into three categories shown as first three rows in \textit{Methods}. The time taken by users through \name{} to achieve comparable accuracies on CIFAR10 and CIFAR100 datasets is less than existing fully automated techniques. This shows the importance of HIL and VA in reducing the resource footprint of NAS. The experiments took approximately 3.5 hrs on \name{} with 8 Tesla V100 GPUs.}
    \begin{tabular}{p{12.045em}p{6.955em}p{5.68em}p{9.09em}p{11.045em}}
    \toprule
    \textbf{Method} & \textbf{GPU days range} & \textbf{Params (MB) range} & \textbf{Error on CIFAR 10} & \textbf{Error on CIFAR 100} \\
    \midrule
    Human Experts & -     & 24-30 & 3.38 $\pm$ 0.8 & 21.76 $\pm$ 6.14 \\
    \midrule
    NAS $>$ 100 GPU days & 150-2000 & 3.3-10.6 & 3.52 $\pm$ 0.34 & 21.62 $\pm$ 6.34\\
    \midrule
    NAS $<$ 5 GPU days & 0.84-50 & 3.3-5.7 & 3.25 $\pm$ 0.8 & 18.5 $\pm$ 2.7 \\
    \midrule
    \name{} (No VA) (T=1K) & 1.4-2.0 & 3.3 & 2.6 $\pm$ 0.1 & 17.8 $\pm$ 0.3 \\
    \midrule
    \textbf{\name{} (Experts)} & \textbf{0.6-0.9} & \textbf{3.1 - 3.7} & \textbf{2.83 $\pm$ 0.2} & \textbf{17.82 $\pm$ 0.7} \\
    \midrule
    \textbf{\name{} (Non-Experts)} & 0.8-1.5 & 2.9-4 & 3.2 $\pm$ 0.3 & 18.6 $\pm$ 0.5 \\
    \bottomrule
    \end{tabular}%
  \label{tab:study2}%
\end{table*}%

\subsection{\name{} architecture search on Imagenet dataset}
In this section, we discuss the architecture search process followed by one of the experts in the study (E1) for architecture search on Imagenet~\cite{krizhevsky2012imagenet}. E1 was first given a short demonstration of using our interface followed by an explanation of the task to be performed. All search steps performed by E1 were logged along with the time taken for each search step. E1 started by loading a customized AlexNet into the interface, which has multiple options of blocks to choose from and follows the basic architecture model of AlexNet. This customized version of AlexNet contains the same number of layers as the original network but each layer has multiple blocks. For example, layer 1 has multiple convolutional blocks, i.e. Conv 3x3, Conv 5x5, and Conv7x7, and similarly for other layers. E1 explained that the reason for choosing the AlexNet template as a template network was his experience in using AlexNet for image classification tasks. After a careful understanding of the template AlexNet, E1 started by analyzing the results of the first 20 iterations of the evolutionary search algorithm, which gives a fitness value for each block of the template network. After the search results from the first four iterations, E1 decided to further analyze the 7x7 Conv blocks from Layers 1 and 2 because the fitness values of these blocks dropped to zero. Focusing on the search space view, E1 was able to find a subspace where the most common block was a7x7 Conv block in Layer1. E1 then dragged this region on the search space view which forced the search algorithm to sample candidate architectures from this search space. After 5 more search iterations, it was confirmed that the presence of this block resulted in a below-par performance of the neural networks; E1 decided to remove this block from the search space using the lego view and then continued the search further. Another 5 iterations of the evolutionary search suggested the removal of 3x3 convolution from Layer 1 and 5x5 convolution from Layer 2; these blocks were removed by E1. Additionally, the search results also suggested that 7x7 convolution was the best at Layer 3 on the evaluation dataset but E1 wasn’t convinced about this result because of his experience. Hence, E1 selected a region in the search space view where the most common block was the 7x7 Conv block at Layer 3 to evaluate more candidate neural networks from this subspace. It was confirmed after a single search iteration that most neural networks from this search space had high accuracy, hence, giving further evidence that 7x7 convolution was the best option among the other blocks at layer 3. The search also suggested that the linear block with 2,304 input and 4,096 output parameters worked the best at layer 6. This yielded the final architecture of the suggested neural network.

\textbf{Results:} E1 compared the results of the suggested network with a baseline of AlexNet performance on the Imagenet dataset after training for 10 epochs. While the baseline AlexNet has an accuracy of 72.70\% on the test data, the network derived from our interface had an accuracy of 74.72\%. This accuracy was further improved to 76.34\% after E1 used his expertise and added batch normalization layer after every convolutional layer in the network. E1 was satisfied with the final network since it resulted in better accuracy than the baseline AlexNet model. This study confirmed that our tool can help computer vision researchers effectively search for and identify high-performing convolutional neural network architectures.

\subsection{Interface Evaluation Through User Study}

\begin{figure}[th]
 \centering
 \includegraphics[width=\columnwidth]{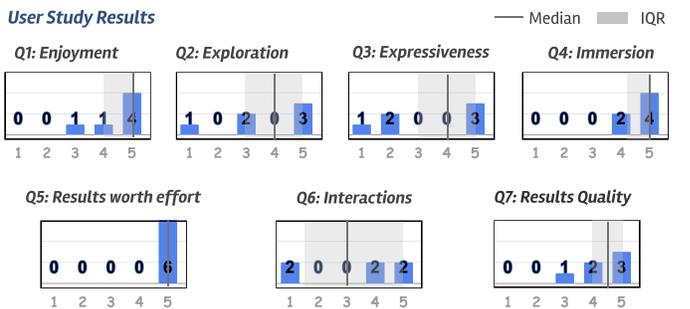}
 \caption{The user study results to evaluate our framework with the Creative Support Index~\cite{cherry2014quantifying}, on a five-point Likert scale. We asked 7 questions to the participants, covering different aspects of usability.}
 \label{fig:likert}
\end{figure}

\label{ss:user_study}
In this study, we evaluated \name{} for its support for multiple factors of usability and creativity. Considering that the main goal of creating \name{} was to add VA and HIL in NAS, we carefully evaluated our interface for its ability to support user thought processes and creativity. Since there are no existing frameworks publicly available that can be used as a suitable baseline for this task (see Table~\ref{tab:comp_hyp}), this user study helped us to explore the strengths and weaknesses of our interface through user feedback.

\textbf{Participants. }
The same participants, as described in Section \ref{ss:case_study} performed this user study. A total of six people participated, out of which three were considered Experts and three were non-experts. 

\textbf{Task and Procedure. }
The task was to answer questions regarding the procedure they applied in the Case Study discussed in Section~\ref{ss:case_study} and rate their experience on a five-point Likert Scale from 1 (Strongly Disagree) to 5 (Strongly Agree). 

The questionnaire was based on 7 factors, 5 of which are taken from the work by Cherry et al.~\cite{cherry2014quantifying} for quantifying the creativity support for design tools (Q1 to Q5 in Figure \ref{fig:likert}). We added two additional factors in the questionnaire to rate the domain-specific questions (Q6 and Q7 in Figure \ref{fig:likert}). The whole study took about 45 minutes for each participant. 

\subsubsection{Questionnaire Results}
As shown in Figure \ref{fig:likert}, out of the total 42 ratings on 7 questions, only 4 ratings had a low score of 1 or 2. Overall, 81\% of the votes rated the questions with a 4 or 5, with Q4 and Q5 being the highest-rated questions with the most Agree votes. Overall, positive feedback and high ratings for design and usability questions show the efficacy of our interface in supporting user creativity in NAS. However, detailed feedback was collected from the experts about the low-scoring questions and some directions of possible improvement in \name{}, discussed in the following text. 

\subsubsection{Expert Interview Results}

Besides the general results, we separately collected detailed feedback from two of the experts (E2 and E3) about their experience in working with \name{}. This interview helped compare \name{} with the existing works from an expert's perspective and helped us gather deeper insights into the VA aspect of our framework. We discuss our results organized by the themes of the questionnaire in the following text.

\textbf{Enjoyment (Q1). }Both the participants found \name{} to be useful in their tasks. E2 explained \textit{``Drag and drop on the template NN was a great way to edit the search and get desired results"}. E3 added \textit{``I like that there are template NN provided for major DL tasks, which we can directly load on the interface and start playing with them."}

\textbf{Exploration (Q2). }Both the participants liked the search space view to exploring the candidate NN search space. E2 commented \textit{``Search Space view is a great idea and the fact that we can see clusters in the candidate NN search space shows how the structures of candidate NN can change the performance of the NNs."} E3 suggested an improvement to \name{} commenting \textit{``A useful feature can be to suggest changes in the NN model based on current fitness scores for some non-experts  or in case the user has no prior idea on what NN will perform better on a given dataset."}

\textbf{Expressiveness (Q3). }Both the experts suggested improvements in the expressiveness aspects of \name{}. E2 suggested  \textit{``the user has great power to do pretty much any change with the interface, which can be great if they know what they are doing. However, in many cases, this can be a disaster if the user mistakenly updates something which later turned out to be useful."} Adding on to this, E3 suggested, \textit{``it's a good idea for future will be to see the impact of user changes on the search results compared to the fully automated EA algorithm. This will allow the users to know the impact of their decision and will make the process more transparent." }

\textbf{Immersion (Q4). }The participants were positive about the immersion part, suggesting a few improvements on top of the existing interactions. E2 commented \textit{``NAS processes are slower than the common design tasks, which are more commonly done through dashboards. Hence, the users cannot see immediate results of their actions in this case."} E3 added to this comment and mentioned \textit{``While the users are waiting for the search iteration to complete, a notification will be useful to see if a particular search iteration has found something useful which can have some impact on the NN performance. Every search result, if it can be linked with user action and its impact, will be useful information to have on the interface."}

\textbf{Results worth effort (Q5). }Both the participants agreed that even small interactions if done correctly, can have a great impact on the search convergence time and the final NN performance. 

\textbf{Interactions  (Q6). }The participants had mixed reactions to the interaction effects of \name{}. E2 commented \textit{``It would be great if we could track and visualize how the scores have changed over the search iterations. It'll further add to the decision-making as we will be able to see the history of changing scores and not just the last timestamp."} E3 was satisfied with the interactions in \name{} and commented \textit{``The search space interactions are useful in controlling the areas to search from. This is a great idea and the fact that I can see all the candidate NNs in a single space is very useful."}

\textbf{Results Quality (Q7). }Both the participants were satisfied with the quality of the results. An improvement was suggested by E2 who said \textit{``Sometimes the clusters in the search space view scatterplot are not very clear, maybe adding supporting plots to show further details of each network architecture would be useful."}

\section{Discussion}
Several important lessons were learned while designing this framework. Our initial discussion with domain experts was decisive in pinning down the main interface design. After all, tasks were formulated within comprehensive discussions with the experts, it was easier to design the visual interface and its components. Also, we realized that adding strong user interaction facilities was important, as a means to allow users to infuse their domain knowledge into the search process to accelerate convergence to the final solution. This design allows analysts to use their domain knowledge and the one-shot search results to quickly converge to the best performing neural network architecture for a given task. Analysts also have the freedom to apply certain soft constraints at their discretion, for example trading off between neural network size and accuracy, for example, different Resnet~\cite{he2016deep} sizes.

\section{Conclusion}
\label{s:conclusion}
This paper presents a visual analytics framework to assist in deep neural network architecture search. Our interface combines the automated one-shot neural network architecture search approach with a human-in-the-loop design. Our interface is also less resource-intensive than conventional automatic neural network architecture search algorithms. Analysts can quickly load a template neural network along with their dataset and explore different subset neural network architectures to find the best one. Our evolutionary search algorithm allows for quick sampling of well-performing candidate architectures which can then be further evaluated for their performance. A design study was conducted in collaboration with several researchers working in the deep learning domain to lay down the tasks to be performed by our interface. We evaluated our framework for its ability to better search for the best performing neural network architecture with the help of a user study, case studies, and expert interviews.

However, besides the effectiveness of our present interface, there remains some scope for improvement, which will be taken on in future work for this project. First, we would like to run comparison experiments to compare the performance of our EA algorithm against other solutions including reinforcement learning and bayesian optimization. Taking some points from our interview evaluation, we will add supporting visualization to the search space view to better present the clusters and candidate NN architectures inside these clusters. We will work on identifying user actions for each search iteration result, that will predict and suggest these actions to the non-expert users. Also, tracking of search iteration results and changes in the fitness scores will be added. We would also like to evaluate \name{} on language models to extend its applicability. Currently, all our evaluation is based on computer vision networks, but as more NAS algorithms start to come from the language domain, it will be interesting to see how \name{} compares against the state-of-the-art transformer models. These features are not yet supported and we will continue to work on our interface to incorporate them in the future. We plan to deploy \name{} for real users as a long-term study to collect in-depth feedback and usage scenarios. 

\section{Acknowledgements}
We would like to thank the anonymous VIS 2022 reviewers for their valuable comments. This work was partially funded by NSF grants CNS 1900706, IIS 1527200 and 1941613, and NSF SBIR contract 1926949.

\bibliographystyle{abbrv-doi}
\bibliography{template}
\end{document}